# Implementation and Comparison of Solution Methods for Decision Processes with Non-Markovian Rewards

**Charles Gretton, David Price, and Sylvie Thiébaux**
Computer Sciences Laboratory
The Australian National University
Canberra, ACT, Australia
{charlesg,davidp,thiebaux}@csl.anu.edu.au

## Abstract

This paper examines a number of solution methods for decision processes with non-Markovian rewards (NMRDPs). They all exploit a temporal logic specification of the reward function to automatically translate the NMRDP into an equivalent Markov decision process (MDP) amenable to well-known MDP solution methods. They differ however in the representation of the target MDP and the class of MDP solution methods to which they are suited. As a result, they adopt different temporal logics and different translations. Unfortunately, no implementation of these methods nor experimental let alone comparative results have ever been reported. This paper is the first step towards filling this gap. We describe an integrated system for solving NMRDPs which implements these methods and several variants under a common interface; we use it to compare the various approaches and identify certain problem features favouring one over the other.

## 1 INTRODUCTION

A decision process in which rewards depend on the sequence of states passed through rather than merely on the current state is called a decision process with *non-Markovian rewards* (NMRDP). In decision-theoretic planning, where many desirable behaviours are more naturally expressed as properties of execution sequences rather than as properties of states, NMRDPs form a more natural model than the commonly adopted fully Markovian decision process (MDP) model [Haddawy and Hanks, 1992; Bacchus et al., 1996].

The more tractable solution methods developed for MDPs do not directly apply to NMRDPs. However, a number of solution methods for NMRDPs have been proposed in the literature [Bacchus et al., 1996; Bacchus et al., 1997; Thiébaux et al., 2002]. These all start with a temporal logic specification of the non-Markovian reward function, which they exploit to automatically translate the NMRDP into an equivalent MDP which is solved using efficient MDP solution methods. The states of this MDP result from augmenting those of the original NMRDP with extra information capturing enough history to make the reward Markovian.

Naturally, there is a tradeoff between the effort spent in the translation, e.g. in producing a *small* equivalent MDP without many irrelevant history distinctions, and the effort required to solve it. Appropriate resolution of this tradeoff depends on the type of representations and solution methods envisioned for the MDP. For instance, *structured* representations and solution methods which have some ability to ignore irrelevant information may cope with a crude translation, while *state-based* (flat) representations and methods will require a more sophisticated translation producing an MDP as small as feasible.

While the approaches [Bacchus et al., 1996; Bacchus et al., 1997; Thiébaux et al., 2002] are all based on translation into an equivalent MDP, they target different MDP representations and solution methods. Specifically, [Bacchus et al., 1996] targets state-based representations and classical solution methods such as value or policy iteration [Howard, 1960]. [Thiébaux et al., 2002] also considers state-based representation but targets heuristic search methods such as LAO* [Hansen and Zilberstein, 2001] or labelled RTDP [Bonet and Geffner, 2003]. Finally, [Bacchus et al., 1997] considers structured representations and solution methods such as structured policy iteration or SPUDD [Boutilier et al., 1995; Hoey et al., 1999].

These different targets lead the three approaches to resolve the translation/solution tradeoff differently, and in turn, to adopt different temporal logics, as appropriate. For instance, both [Bacchus et al., 1996; Bacchus et al., 1997] use *linear temporal logic* with *past* operators (PLTL), as this yields a straightforward semantics of non-Markovian rewards, and lends itself to a simple characterisation of a range of translations, from the crudest to the finest. [Thiébaux et al., 2002], on the other hand, relies on a more complex extension of LTL with *future* operators ($FLTL), as it naturally leads to a style of translation suited to the needs of heuristic search methods.

The three cited papers agree that the most important item for future work is the implementation and experimental comparison of the respective approaches, with a view to identifying the features that favour one over the other. This paper is the first step in that direction. We start with a review of NMRDPs and of the three approaches. We then describe NMRDPP (NMRDP Planner), an integrated system which implements, under a single interface, a family of NMRDP solution methods based on the cited approaches, and reports a range of statistics about their performance.

We use this system to compare their behaviours under the influence of various factors such as the structure and degree of uncertainty in the dynamics, the class of rewards and the syntax used to describe them, reachability, and relevance of rewards to the optimal policy.

## 2 NMRDP SOLUTION METHODS

### 2.1 MDPs, NMRDPs, EQUIVALENCE

We start with some notation and definitions. Given a finite set $S$ of states, we write $S^*$ for the set of finite sequences of states over $S$, and $S^\omega$ for the set of possibly infinite state sequences. Where '$\Gamma$' stands for a possibly infinite state sequence in $S^\omega$ and $i$ is a natural number, by '$\Gamma_i$' we mean the state of index $i$ in $\Gamma$, and by '$\Gamma(i)$' we mean the prefix $\langle \Gamma_0, \ldots, \Gamma_i \rangle \in S^*$ of $\Gamma$.

We take a Markov decision process to be a 5-tuple $\langle S, s_0, A, \Pr, R \rangle$, where $S$ is a finite set of fully observable states, $s_0 \in S$ is the initial state, $A$ is a finite set of actions ($A(s)$ denotes the subset of actions applicable in $s \in S$), $\{\Pr(s, a, \bullet) \mid s \in S, a \in A(s)\}$ is a family of probability distributions over $S$ such that $\Pr(s, a, s')$ is the probability of being in state $s'$ after performing action $a$ in state $s$, and $R : S \mapsto \mathbf{R}$ is a reward function such that $R(s)$ is the immediate reward for being in state $s$. A stationary policy for an MDP is a function $\pi : S \mapsto A$, such that $\pi(s) \in A(s)$ is the action to be executed in state $S$. The value $V(\pi)$ of the policy, which we seek to maximise, is the sum of the expected future rewards, discounted by how far into the future they occur:

$$V(\pi) = \lim_{n \to \infty} \mathsf{E}\left[\sum_{i=0}^{n} \beta^i R(\Gamma_i) \mid \pi, \Gamma_0 = s_0\right]$$

where $0 \leq \beta \leq 1$ is the discounting factor controlling the contribution of distant rewards.

A decision process with non-Markovian rewards is identical to an MDP except that the domain of the reward function is $S^*$. The idea is that if the process has passed through state sequence $\Gamma(i)$ up to stage $i$, then the reward $R(\Gamma(i))$ is received at stage $i$. Like the reward function, a policy for an NMRDP depends on history, and is a mapping from $S^*$ to $A$ with $\pi(\Gamma(i)) \in A(\Gamma_i)$. As before, the value of $\pi$ is the expectation of the discounted cumulative reward over an infinite horizon:

$$V(\pi) = \lim_{n \to \infty} \mathsf{E}\left[\sum_{i=0}^{n} \beta^i R(\Gamma(i)) \mid \pi, \Gamma_0 = s_0\right]$$

The solution methods considered here operate by translating an NMRDP into an equivalent MDP with an extended state space [Bacchus et al., 1996]. The states in this MDP, which, for clarity, we will sometimes call expanded states (*e-states*, for short), augment the states of the NMRDP by encoding additional information sufficient to make the reward history-independent. For instance, if we only want to reward the very first achievement of goal $g$ in an NMRDP, the states of an equivalent MDP would carry at most one extra bit of information recording whether $g$ has already been true. In the following, we see an e-state as labelled by a state of the NMRDP (via the function $\tau$ below) and by historical information. The dynamics of NMRDPs being Markovian, the actions and their probabilistic effects in the MDP are exactly those of the NMRDP.

Formally, MDP $D' = \langle S', s_0', A', \Pr', R' \rangle$ is equivalent to NMRDP $D = \langle S, s_0, A, \Pr, R \rangle$ if there exists a mapping $\tau : S' \mapsto S$ such that:

1. $\tau(s_0') = s_0$.
2. For all $s' \in S'$, $A'(s') = A(\tau(s'))$.
3. For all $s_1, s_2 \in S$, if there is $a \in A(s_1)$ such that $\Pr(s_1, a, s_2) > 0$, then for all $s_1' \in S'$ such that $\tau(s_1') = s_1$, there exists a unique $s_2' \in S'$, $\tau(s_2') = s_2$, such that for all $a \in A'(s_1')$, $\Pr'(s_1', a, s_2') = \Pr(s_1, a, s_2)$.
4. For any feasible[1] state sequence $\Gamma$ for $D$ and any feasible state sequence $\Gamma'$ for $D'$ such that $\Gamma_0 = s_0$ and $\forall i \; \tau(\Gamma_i') = \Gamma_i$, we have: $\forall i \; R'(\Gamma_i') = R(\Gamma(i))$.

Items 1–3 ensure that there is a bijection between feasible state sequences in the NMRDP and feasible e-state sequences in the equivalent MDP. Therefore, a stationary policy for the MDP can be reinterpreted as a non-stationary policy for the NMRDP. Furthermore, item 4 ensures that the two policies have identical values, and that consequently, solving an NMRDP optimally reduces to producing an equivalent MDP and solving it optimally [Bacchus et al., 1996].

When solving NMRDPs in this setting, the central issue is to choose a language for compactly representing non-Markovian reward functions and a translation algorithm which is *adapted* to the needs of the MDP representations and solution methods we are targeting. In particular, this choice should enable an appropriate resolution of the trade-off between the time spent in the translation and the time spent in solving the resulting MDP. The three approaches we consider have different targets, for which different languages and translations are appropriate. We now present the main ideas behind these approaches. For details, the reader is referred to the respective papers.

### 2.2 PLTLSIMP AND PLTLMIN

[Bacchus et al., 1996] targets state-based MDP representations. The equivalent MDP is first entirely generated—this involves the enumeration of all its states and transitions. Then, it is solved using dynamic programming methods such as value or policy iteration. Since these methods are very sensitive to the number of states, attention is paid to producing a minimal equivalent MDP (with the least number of states).

The language chosen to represent rewards is a linear temporal logic of the past (PLTL). The syntax of PLTL is that of propositional logic, augmented with the operators $\ominus$ (previously) and $\mathsf{S}$ (since). Whereas a classical propositional logic formula denotes a set of states (a subset of $S$), a PLTL formula denotes a set of finite *sequences* of states (a subset of $S^*$). A formula without temporal modality expresses a property that must be true of the current state, i.e., the last state of the finite sequence. $\ominus \phi$ specifies that $\phi$ holds in the previous state (the state one before the last). We will write $\ominus^k$ ($k$ times ago), for $k$ iterations of the $\ominus$ modality. $\phi_1 \mathsf{S} \phi_2$, requires $\phi_2$ to have been true at some point in the sequence, and $\phi_1$ to have held since right after then. From S, one can define the useful operators $\Diamond f \equiv \top \mathsf{S} f$ meaning that $f$ has been true at some point, and $\boxminus f \equiv \neg \Diamond \neg f$ meaning that $f$ has always been true. E.g, $g \wedge \neg \ominus \Diamond g$ denotes the set of finite sequences ending in a state where $g$ is true for the first time in the sequence. We describe reward functions with a set of pairs $\phi_i : r_i$ where $\phi_i$ is a PLTL reward formula and $r_i$ is a real, with the semantics that the reward assigned to a sequence in $S^*$ is the sum of the $r_i$s for which that sequence is a model of $\phi_i$. Below, we let $\Phi$ denote the set of reward formulae $\phi_i$ in the description of the reward function.

---

[1] All transitions along the sequence have non-zero probability.

The translation into an MDP relies on the equivalence $\phi_1 \mathsf{S} \phi_2 \equiv \phi_2 \vee (\phi_1 \wedge \ominus(\phi_1 \mathsf{S} \phi_2))$, with which we can decompose temporal modalities into a requirement about the last state $\Gamma_i$ of a sequence $\Gamma(i)$, and a requirement about the prefix $\Gamma(i-1)$ of the sequence. More precisely, given state $s$ and a given formula $\phi$, one can compute in[2] $\mathcal{O}(||\Phi||)$ a new formula $\text{Reg}(\phi, s)$ called the regression of $\phi$ through $s$. Regression has the property that $\phi$ is true of a finite sequence $\Gamma(i)$ ending with $\Gamma_i = s$ iff $\text{Reg}(\phi, s)$ is true of the prefix $\Gamma(i-1)$. That is, $\text{Reg}(\phi, s)$ represents what must have been true previously for $\phi$ to be true now.

The translation exploits the PLTL representation of rewards as follows. Each e-state in the generated MDP is labelled with a set $\Psi \subseteq \overline{\text{Sub}(\Phi)}$ of subformulae of the reward formulae in $\Phi$ (and their negations).[3] The subformulae in $\Psi$ must be (1) true of the paths leading to the e-state, and (2) sufficient to determine the current truth of all reward formulae in $\Phi$, as this is needed to compute the current reward. Ideally the $\Psi$s should also be (3) small enough to enable just that, i.e. they should not contain subformulae which draw history distinctions which are irrelevant to determining the reward at one point or another. Note however that in the worst-case, the number of distinctions needed, even in the minimal equivalent MDP, may be exponential in $||\Phi||$. This happens for instance with the formula $\ominus^k \phi$, which requires $k$ additional bits of information memorising the truth of $\phi$ over the last $k$ steps.

For the choice of the $\Psi$s, [Bacchus et al., 1996] considers two cases. In the simple case, which we call PLTLSIM, an MDP obeying properties (1) and (2) is produced by simply labelling each e-state with the set of *all* subformulae in $\overline{\text{Sub}(\Phi)}$ which are true of the sequence leading to that e-state. This MDP is generated forward, starting from the initial e-state labelled with $s_0$ and with the set $\Psi_0 \subseteq \overline{\text{Sub}(\Phi)}$ of all subformulae which are true of the sequence $(s_0)$. The successors of any e-state labelled by NMRDP state $s$ and subformula set $\Psi$ are generated as follows: each of them is labelled by a successor $s'$ of $s$ in the NMRDP and by the set of subformulae $\{\psi' \in \overline{\text{Sub}(\Phi)} \mid \Psi \models \text{Reg}(\psi', s')\}$.

Unfortunately, this simple MDP is far from minimal. Although it could be postprocessed for minimisation before the MDP solution method is invoked, the above expansion may still constitute a serious bottleneck. Therefore, [Bacchus et al., 1996] considers a more complex two-phase translation, which we call PLTLMIN, capable of producing an MDP also satisfying property (3). Here, a preprocessing phase iterates over all states in $S$, and computes, for each state $s$, a set $l(s)$ of subformulae, where the function $l$ is the solution of the fixpoint equation $l(s) = \{\Phi \cup \{\text{Reg}(\psi', s')\} \mid \psi' \in l(s'), s' \text{ is a successor of } s\}$. Only subformulae in $\overline{l(s)}$ will be candidates for inclusion in the sets labelling the respective e-states labelled with $s$. That is, the subsequent expansion phase will be as above, but taking $\Psi_0 \subseteq \overline{l(s_0)}$ and $\psi' \subseteq \overline{l(s')}$ instead of $\Psi_0 \subseteq \overline{\text{Sub}(\Phi)}$ and $\psi' \subseteq \overline{\text{Sub}(\Phi)}$. As the subformulae in $l(s)$ are exactly those that are relevant to the way actual execution sequences starting from e-states labelled with $s$ are rewarded, this leads the expansion phase to produce a minimal equivalent MDP.

In the worst case, computing this $l$ requires a space, and a number of iterations through $S$, exponential in $||\Phi||$. Hence the question arises of whether the gain during the expansion phase is worth the extra complexity of the preprocessing phase. This is one of the questions we will try to answer.

### 2.3 PLTLSTRUCT

The approach in [Bacchus et al., 1997], which we call PLTLSTR, targets structured MDP representations: the transition model, policies, reward and value functions are represented in a compact form, e.g. as trees or algebraic decision diagrams (ADDs). [Boutilier et al., 1995; Hoey et al., 1999]. For instance, the probability of a given proposition (state variable) being true after the execution of an action is specified by a tree whose leaves are labelled with probabilities, whose nodes are labelled with the state variables on whose previous values the given variable depends, and whose arcs are labelled by the possible previous values ($\top$ or $\bot$) of these variables. The translation amounts to augmenting the compact representation of the transition model with new *temporal* variables together with the compact representation of (1) their dynamics, e.g. as a tree over the previous values of the relevant variables, and (2) of the non-Markovian reward function in terms of the variables' current values. Then, structured solution methods such as structured policy iteration or the SPUDD algorithm are run on the resulting structured MDP. Neither the translation nor the solution methods explicitly enumerates states.

The PLTLSTR translation can be seen as a symbolic version of PLTLSIM. The set $T$ of added temporal variables contains the purely temporal subformulae PTSub($\Phi$) of the reward formulae in $\Phi$, to which the $\ominus$ modality is prepended (unless already there): $T = \{\ominus\psi \mid \psi \in \text{PTSub}(\Phi), \psi \neq \ominus\psi'\} \cup \{\ominus\psi \mid \ominus\psi \in \text{PTSub}(\Phi)\}$. Thus, by repeatedly applying the equivalence $\phi_1 \mathsf{S} \phi_2 \equiv \phi_2 \vee (\phi_1 \wedge \ominus(\phi_1 \mathsf{S} \phi_2))$ to any subformula in PTSub($\Phi$), we can express its current value, and hence that of reward formulae, as a function

---

[2]The size $||\Phi||$ of a set of reward formulae $\Phi$ is measured as the sum of the lengths of the formulae in $\Phi$.

[3]Given a set $F$ of formulae, we write $\overline{F}$ for $F \cup \{\neg f \mid f \in F\}$

of the current values of formulae in $T$ and state variables, as required by the compact representation of the transition model.

As with PLTLSIM, the underlying MDP is far from minimal—the encoded history features do not even vary from one state to the next. However, size is not as problematic as with state-based approaches, because structured solution methods do not enumerate states and are able to dynamically ignore some of the variables that become irrelevant during policy construction. For instance, when solving the MDP, they may be able to determine that some temporal variables have become irrelevant because the situation they track, although possible in principle, is too risky to be realised under a good policy. This *dynamic* analysis of rewards contrasts with the *static* analysis in [Bacchus et al., 1996] which must encode enough history to determine the reward at all reachable futures under any policy.

One question that arises is that of the circumstances under which this analysis of irrelevance by structured solution methods, especially the dynamic aspects, is really effective. This is another question this paper will try to address.

### 2.4 FLTL

The approach in [Thiébaux et al., 2002], which we call FLTL, considers state-based representations of the equivalent MDP and targets heuristic forward search solution methods such as LAO* or labelled RTDP. Starting from a compact representation of the MDP and an admissible heuristic, these methods need only explicitly generate and explore a fraction of the state space to produce an optimal solution. To gain maximum benefit from these methods, the translation into MDP must avoid generating states and e-states that the method would not generate. Therefore, the FLTL translation operates entirely on-line: the solution method is given full control of which parts of the MDP are generated and explored. This contrasts with PLTLMIN, which requires an off-line preprocessing phase iterating through all states in $S$.

[Thiébaux et al., 2002] notes that when using PLTL to specify rewards, there does not seem to be a way of designing an on-line translation producing an MDP of acceptable size.[4] Instead, [Thiébaux et al., 2002] adopts a variant of LTL with *future* operators called $FLTL. The syntax is that of negation normal form propositional logic augmented with the constant $ (rewarded) and the operators $\bigcirc$ (next) and $U$ (weak until). As in PLTL, a $FLTL formula represents a subset of $S^*$ – see [Thiébaux et al., 2002] for a formal semantics [5]. But given the forward looking character of the language, it is best to see a formula as a recipe for distributing rewards, starting from the current state (i.e., the first state of the rest of the sequence). Informally, $ means that we get rewarded now. $\bigcirc \phi$ means that $\phi$ holds in the next state, and $\phi_1 U \phi_2$ means that $\phi_1$ will be true from now on until $\phi_2$ becomes true, if ever. From $U$, one can define $\Box \phi \equiv \phi U \bot$, meaning that $\phi$ will always be true. E.g, $\Box(c \to \Box(\phi \to \Box\$))$ means that following a command $c$, we will be rewarded from the moment $\phi$ holds onwards. $\neg \phi U (\phi \wedge \$)$ means that we will be rewarded the first time $\phi$ becomes true. As with PLTL, a reward function is represented by a set of pairs consisting of a formula and a real.

The translation is based on a variant of progression [Bacchus and Kabanza, 2000], which is to future-oriented logics what regression is to past-oriented ones: $Prog(\phi, s)$ tells us what must hold next for $\phi$ to hold now, at the current state $s$. Each e-state in the equivalent MDP is labelled by a state of the NMRDP and by a set of $FLTL formulae. The initial e-state is labelled with $s_0$ and the set $\Phi_0$ of all reward formulae in the given reward function. Each successor of an e-state labelled with $s$ and $\Phi$ is labelled by a successor $s'$ of $s$ in the NMRDP and by the set $\{\$Prog(\phi, s) \mid \phi \in \Phi\}$ of the progressions of the formulae in $\Phi$ through $s$. Although the MDP produced that way is not minimal, it satisfies a weaker but still interesting notion of minimality, called blind minimality. Intuitively, a blind minimal equivalent MDP is the smallest equivalent MDP achievable by any on-line translation.

With FLTL, the structure of the reward formulae is preserved by the translation and exploited by progression. This contrasts with PLTLSIM which completely loses this structure by considering subformulae individually. One of the purposes of the preprocessing phase in PLTLMIN is to recover this structure. One question that arises is whether the simplicity of the FLTL translation combined with the power of heuristic search compensates for the weakness of blind minimality, or whether the benefits of true minimality as in PLTLMIN outweigh the cost of the preprocessing phase. Furthermore, with FLTL, as with PLTLSTR, the analysis of rewards is performed dynamically, as a function of how the search proceeds. Another question we will try to answer is whether the respective dynamic analyses are equally powerful.

## 3 THE NMRDP PLANNER

The first step towards a decent comparison of the different approaches is to have a framework that includes them all. The non-Markovian reward decision process planner[6], NMRDPP, provides an implementation of the approaches in a common framework, within a single system, and with a common input language.

The input language enables the specification of actions, initial states, rewards, and control-knowledge. The format for the action specification is essentially the same as in the SPUDD system [Hoey et al., 1999]. When the input is parsed, the action specification trees are converted into ADDs by the CUDD package. The reward specification is one or more formulae, each associated with a real. These formulae are in either PLTL or $FLTL and are stored as trees by the system. Control knowledge is given in the same language as that chosen for the reward. Control knowledge formulae will have to be verified by any sequence of states feasible under the generated policies. Initial states are simply specified as part of the control knowledge or as explicit assignments to propositions.

---

[4] PLTLSIM can be performed entirely on-line, but leads to a large MDP.

[5] This is more complex than the standard FLTL semantics. The interpretation of $ is not fixed: $ is made true only when needed to ensure that the formula holds (in the classical FLTL sense of the term) in every sequence of $S^\omega$. For reasons of readability and space, the text above is deliberately evasive.

[6] http://discus.anu.edu.au/~charlesg/nmrdpp

The common framework underlying NMRDPP takes advantage of the fact that NMRDP solution methods can, in general, be divided into the distinct phases of preprocessing, expansion, and solving. The first two are optional. For PLTLSIM, *preprocessing* computes the set $\overline{\mathrm{Sub}(\Phi)}$ of subformulae of the reward formulae. For PLTLMIN, it also includes computing the labels $l(s)$ for each state $s$. For PLTLSTR, it involves computing the set $T$ of temporal variables as well as the ADDs for their dynamics and for the rewards. FLTL does not require any preprocessing. *Expansion* is the optional generation of the entire equivalent MDP prior to solving. Whether or not off-line expansion is sensible depends on the MDP solution method used. If state-based value or policy iteration is used, then the MDP needs to be expanded anyway. If, on the other hand, a heuristic search algorithm or structured method is used, it is definitely a bad idea. In our experiments, we often used expansion solely for the purpose of measuring the size of the generated MDP. *Solving* the MDP can be done using a number of methods. Currently, NMRDPP provides implementations of classical dynamic programming methods, namely state-based value and policy iteration [Howard, 1960], of heuristic search methods: state-based LAO* [Hansen and Zilberstein, 2001] using either value or policy iteration as a subroutine, and of one structured method, namely SPUDD [Hoey et al., 1999].

Altogether, the various types of preprocessing, the choice of whether to expand, and the MDP solution methods, give rise to quite a number of NMRDP approaches, including, but not limited to those previously mentioned For instance, we obtain an interesting variant of PLTLSTR, which we call PLTLSTR(A), by considering additional preprocessing, whereby the state space is explored (without explicitly enumerating it) to produce a BDD representation of the e-states reachable from the start state. This is done by starting with a BDD representing the start e-state, and repeatedly applying each action. Non-zero probabilities are converted to ones and the result "or-ed" with the last result. When no action adds any reachable e-states to this BDD, we can be sure it represents the reachable e-state space. This is then used as additional control knowledge to restrict the search. It should be noted that without this phase PLTLSTR makes no assumptions about the start state, thus is left at a possible disadvantage. Similar techniques have been used in the symbolic implementation of LAO* [Feng and Hansen, 2002]. Given temporal variables are also included in the BDD, PLTLSTR(A) is able to exploit reachability in the space of e-states as PLTLMIN does in the state-based case.

NMRDPP is implemented in C++, and makes use of a number of supporting libraries. In particular, the structured algorithms rely heavily on the CUDD library for representing decision diagrams. The non-structured algorithms make use of the MTL—Matrix Template Library for matrix operations. MTL takes advantage of modern processor features such as MMX and SSE and provides efficient sparse matrix operations. We believe that our implementations of MDP solution methods are comparable with the state of the art. For instance, we found that our implementation of SPUDD is comparable in performance (within a factor of 2) to the reference implementation [Hoey et al., 1999].

## 4 EXPERIMENTAL OBSERVATIONS

We are faced with three substantially different approaches which are not easy to compare, as their performance will depend on domain features as varied as the structure in the transition model, reachability, the type, syntax, and length of the temporal reward formula, the availability of good heuristics and control-knowledge, etc, and on the interactions between these factors. In this section, we try to answer the questions raised above and report an experimental investigation into the influence of some of these factors: dynamics, reward type, syntax, reachability, and presence of rewards irrelevant to the optimal policy. In some cases but not all, we were able to identify systematic patterns. The results were obtained using a Pentium4 2.6GHz GNU/Linux 2.4.20 machine with 500MB of ram.

### 4.1 PRELIMINARY REMARKS

Clearly, FLTL and PLTLSTR(A) have great potential for exploiting domain-specific heuristics and control-knowledge; PLTLMIN less so. To avoid obscuring the results, we therefore refrained from incorporating these features in the experiments. When running LAO*, the heuristic value of a state was the crudest possible (the sum of all reward values in the problem). Performance results should be interpreted in this light – they do not necessarily reflect the practical abilities of the methods that can exploit these features.

We begin with some general observations. One question raised above was whether the gain during the expansion phase is worth the expensive preprocessing performed by PLTLMIN, i.e. whether PLTLMIN typically outperforms PLTLSIM. We can definitively answer this question: up to pathological exceptions, preprocessing pays. We found that expansion was the bottleneck, and that post-hoc minimisation of the MDP produced by PLTLSIM did not help much. PLTLSIM is therefore of little or no practical interest, and we decided not to report results on its performance, as it is often an order of magnitude worse than that of PLTLMIN. Unsurprisingly, we also found that PLTLSTR would typically scale to larger state spaces, inevitably leading it to outperform state-based methods. However, this effect is not uniform: structured solution methods sometimes impose excessive memory requirements which makes them uncompetitive in certain cases, for example where $\ominus^n \phi$, for large $n$, features as a reward formula.

### 4.2 DOMAINS

Experiments were performed on four hand-coded domains (propositions + dynamics) and on random domains. Each hand-coded domain has $n$ propositions $p_i$, and a dynamics which makes every state possible and eventually reachable from the initial state in which all propositions are false. The first two such domains, SPUDD-LINEAR and SPUDD-EXPON were discussed in [Hoey et al., 1999]; the two others are our own. The intention of SPUDD-LINEAR was to take advantage of the best case behaviour of SPUDD. For each proposition $p_i$, it has an action $a_i$ which sets $p_i$ to true and all propositions $p_j$, $1 \leq j < i$ to false. SPUDD-EXPON, was used in [Hoey et al., 1999] to demonstrate the worst case behaviour of SPUDD. For each proposition $p_i$, it has an action $a_i$ which sets $p_i$ to true only when all propositions $p_j$, $1 \leq j < i$ are true (and sets $p_i$ to false otherwise), and sets the latter propositions to false. The third domain, called ON/OFF, has one "turn-on" and one "turn-off" action per proposition. The "turn-on-$p_i$" action only probabilistically succeeds in setting $p_i$ to true when $p_i$ was false. The turn-off action is similar. The fourth domain, called COMPLETE, is a fully connected reflexive domain. For each proposition $p_i$ there is an action $a_i$ which sets $p_i$ to true with probability $i/(n+1)$ (and to false otherwise) and $p_j$, $j \neq i$ to true or false with probability 0.5. Note that $a_i$ can cause a transition to any of the $2^n$ states.

Random domains of size $n$ also involve $n$ propositions. The method for generating their dynamics is out of the scope of this paper, but let us just mention that we are able to generate random dynamics exhibiting a given degree of "structure" and a given degree of uncertainty. Lack of structure essentially measures the bushiness of the internal part of the ADDs representing the actions, and uncertainty measures the bushiness of their leaves.

### 4.3 INFLUENCE OF DYNAMICS

The interaction between dynamics and reward certainly affects the performance of the different approaches, though not so strikingly as other factors such as the reward type (see below). We found that under the same reward scheme, varying the degree of structure or uncertainty did not generally change the relative success of the different approaches. For instance, Figures 1 and 2 show the average run time of the methods as a function of the degree of structure, resp. degree of uncertainty, for random problems of size $n = 6$ and reward $\ominus^n \neg \ominus \top$ (the state encountered at stage $n$ is rewarded, regardless of its properties[7]). Run-time increases slightly with both degrees, but there is no significant change in relative performance. These are typical of the graphs we obtain for other rewards.

Clearly, counterexamples to this observation exist. These are most notable in cases of extreme dynamics, for instance with the SPUDD-EXPON domain. Although for small values of $n$, such as $n = 6$, PLTLSTR approaches are faster than the others in handling the reward $\ominus^n \neg \ominus \top$ for virtually any type of dynamics we encountered, they perform very poorly with that reward on SPUDD-EXPON. This is explained by the fact that only a small fraction of SPUDD-EXPON states are reachable in the first $n$ steps. After $n$ steps, FLTL immediately recognises that reward is of no consequence, because the formula has progressed to $\top$. PLTLMIN discovers this fact only after expensive preprocessing. PLTLSTR, on the other hand, remains concerned by the prospect of reward, just as PLTLSIM would.

### 4.4 INFLUENCE OF REWARD TYPES

The type of reward appears to have a stronger influence on performance than dynamics. This is unsurprising, as the reward type significantly affects the size of the generated MDP: certain rewards only make the size of the minimal equivalent MDP increase by a constant number of states or a constant factor, while others make it increase by a factor exponential in the length of the formula. Table 1 illustrates this. The third column reports the size of the minimal equivalent MDP induced by the formulae on the left hand side.[8]

A legitimate question is whether there is a direct correlation between size increase and (in)appropriateness of the different methods. For instance, we might expect the state-based methods to do particularly well in conjunction with reward types inducing a small MDP and otherwise badly in comparison with structured methods. Interestingly, this is not always the case. For instance, in Table 1 whose last two columns report the fastest and slowest methods over the range of hand-coded domains where $1 \leq n \leq 12$, the first row contradicts that expectation. Moreover, although PLTLSTR is fastest in the last row, for larger values of $n$ (not represented in the table), it aborts through lack of memory, unlike the other methods.

The most obvious observations arising out of these experiments is that PLTLSTR is nearly always the fastest—until it runs out of memory. Perhaps the most interesting results are those in the second row, which expose the inability of methods based on PLTL to deal with rewards specified as long sequences of events. In converting the reward formula to a set of subformulae, they lose information about the order of events, which then has to be recovered laboriously by reasoning. $FLTL progression in contrast takes the events one at a time, preserving the relevant structure at each step. Further experimentation led us to observe that all PLTL based algorithms perform poorly where reward is specified using formulae of the form $\ominus^k \phi$, $\vee_{i=1}^{k} \ominus^i \phi$, and $\wedge_{i=1}^{k} \ominus^i \phi$ ($\phi$ has been true $k$ steps ago, within the last $k$ steps, or at all the last $k$ steps).

### 4.5 INFLUENCE OF SYNTAX

Unsurprisingly, we find that the syntax used to express rewards, which affects the length of the formula, has a major influence on the run time. A typical example of this effect

---

[7]$\bigcirc^n \$ in $FLTL

[8]The figures are not necessarily valid for non-completely connected NMRDPs. Unfortunately, even for completely connected domains, there does not appear to be a much cheaper way to determine the MDP size than to generate it and count states.

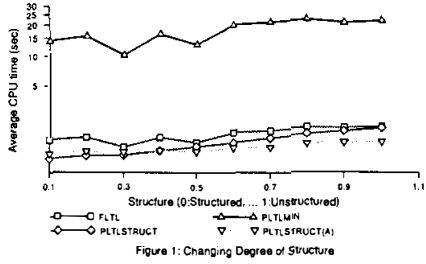
Figure 1: Changing Degree of Structure

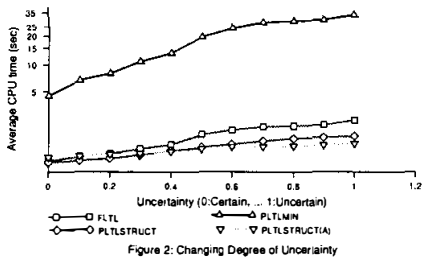
Figure 2: Changing Degree of Uncertainty

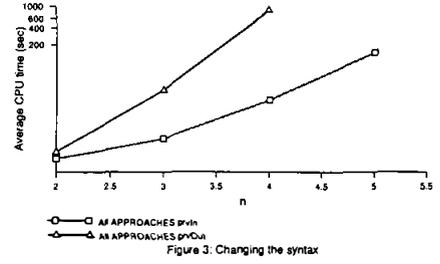
Figure 3: Changing the syntax

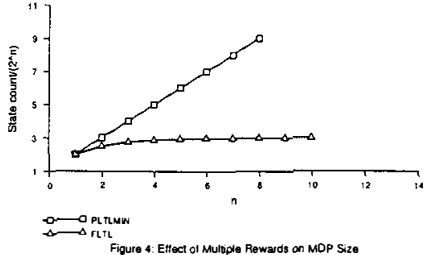
Figure 4: Effect of Multiple Rewards on MDP Size

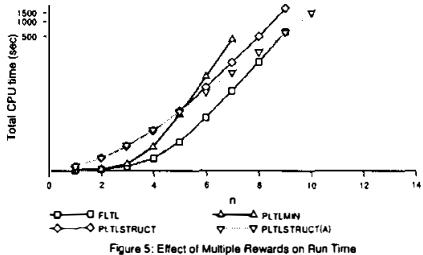
Figure 5: Effect of Multiple Rewards on Run Time

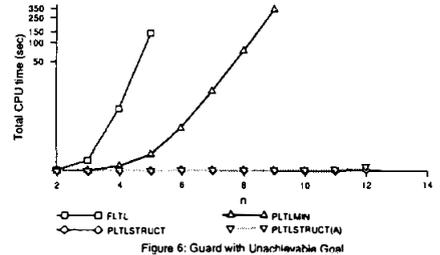
Figure 6: Guard with Unachievable Goal

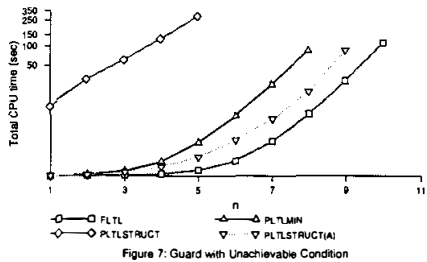
Figure 7: Guard with Unachievable Condition

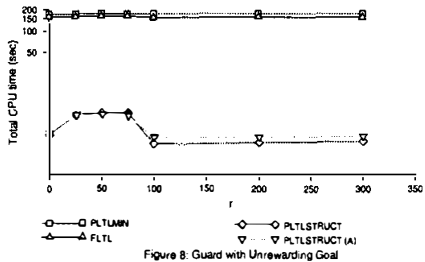
Figure 8: Guard with Unrewarding Goal

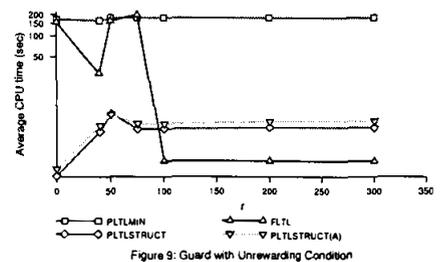
Figure 9: Guard with Unrewarding Condition

Table 1: Influence of reward type on MDP size and method performance

| type | formula | size | fastest | slowest |
|---|---|---|---|---|
| first time all $p_i$s | $(\wedge_{i=1}^n p_i) \wedge (\neg \ominus \diamondsuit \wedge_{i=1}^n p_i)$ | $\mathcal{O}(1)\|\|S\|\|$ | PLTLSTR(A) | PLTLMIN |
| $p_i$s in sequence from start state | $(\wedge_{i=1}^n \ominus^i p_i) \wedge \ominus^n \neg \ominus \top$ | $\mathcal{O}(n)\|\|S\|\|$ | FLTL | PLTLSTR |
| two consecutive $p_i$s | $\vee_{i=1}^{n-1}(\ominus p_i \wedge p_{i+1})$ | $\mathcal{O}(n^k)\|\|S\|\|$ | PLTLSTR | FLTL |
| all $p_i$s $n$ times ago | $\ominus^n \wedge_{i=1}^n p_i$ | $\bullet(2^n)\|\|S\|\|$ | PLTLSTR | PLTLMIN |

is captured in Figure 3. This graph demonstrates how re-expressing $prvOut \equiv \ominus^n(\wedge_{i=1}^n p_i)$ as $prvIn \equiv \wedge_{i=1}^n \ominus^n p_i$, thereby creating $n$ times more temporal subformulae, alters the running time of all PLTL methods. FLTL is affected too as $FLTL progression requires two iterations through the formula. The graph represents the averages of the running times over all the methods, for the COMPLETE domain.

Our most serious concern in relation to the PLTL approaches is their handling of reward specifications containing multiple reward elements. Most notably we found that PLTLMIN does not necessarily produce the minimal equivalent MDP in this situation. To demonstrate, we consider the set of reward formulae $\{\phi_1, \phi_2, \ldots, \phi_n\}$, each associated with the same real value $r$. Given this, PLTL approaches will distinguish unnecessarily between past behaviours which lead to identical future rewards. This may occur when the reward at an e-state is determined by the truth value of $\phi_1 \vee \phi_2$. This formula does not necessarily require e-states that distinguish between the cases in which $\{\phi_1 \equiv \top, \phi_2 \equiv \bot\}$ and $\{\phi_1 \equiv \bot, \phi_2 \equiv \top\}$ hold; however, given the above specification, PLTLMIN shall make this distinction. For example, taking $\phi_i = \ominus p_i$, Figure 4 shows that FLTL leads to an MDP whose size is at most 3 times that of the NMRDP. In contrast, the relative size of the MDP produced by PLTLMIN is linear in $n$, the number of rewards and propositions. These results are obtained with all hand-coded domains except SPUDD-EXPON. Figure 5 shows the run-times as a function of $n$ for COMPLETE. FLTL dominates and is only overtaken by PLTLSTR(A) for large values of $n$, when the MDP becomes too large for explicit exploration to be practical. To obtain the minimal equivalent MDP using PLTLMIN, a bloated reward specification of the form $\{\ominus \vee_{i=1}^n (p_i \wedge_{j=1, j \neq i}^n \neg p_j) : r, \ldots, \ominus \wedge_{i=1}^n p_i : n * r\}$ is necessary, which, by virtue of its exponential length, is not an adequate solution.

### 4.6 INFLUENCE OF REACHABILITY

All approaches claim to have some ability to ignore variables which are irrelevant because the condition they track is unreachable: PLTLMIN detects them through preprocessing, PLTLSTR exploits the ability of structured solution methods to ignore them, and FLTL ignores them when progression never exposes them. However, given that the mechanisms for avoiding irrelevance are so different, we expect corresponding differences in their effects. On experimental investigation, we found that the differences in performance are best illustrated by looking at *guard formulae*, which assert that if a trigger condition $c$ is reached then a reward will be received upon achievement of the goal $g$ in, resp. within, $k$ steps. In PLTL, this is written $g \wedge \ominus^k c$, resp. $g \wedge \vee_{i=1}^k \ominus^i c$, and in \$FLTL, $\Box(c \rightarrow \bigcirc^k(g \rightarrow \$))$, resp. $\Box(c \rightarrow \wedge_{i=1}^k \bigcirc^i(g \rightarrow \$))$.

Where the *goal* $g$ is unreachable, PLTL approaches perform well. As it is always false, $g$ does not lead to behavioural distinctions. On the other hand, while constructing the MDP, FLTL considers the successive progressions of $\bigcirc^k g$ without being able to detect that it is unreachable until it actually fails to happen. This is exactly what the blindness of blind minimality amounts to. Figure 6 illustrates the difference in performance as a function of the number $n$ of propositions involved in the SPUDD-LINEAR domain, when the reward is of the form $g \wedge \ominus^n c$, with $g$ unreachable.

FLTL shines when the *trigger* $c$ is unreachable: the formula will always progress to itself, and the goal, however complicated, is never tracked in the generated MDP. In this situation PLTL approaches still consider $\ominus^k c$ and its subformulae, only to discover, after expensive preprocessing for PLTLMIN, after reachability analysis for PLTLSTR(A), and never for PLTLSTR, that these are irrelevant. This is illustrated in Figure 7, again with SPUDD-LINEAR and a reward of the form $g \wedge \ominus^n c$, with $c$ unreachable.

### 4.7 DYNAMIC IRRELEVANCE

[Bacchus et al., 1997; Thiébaux et al., 2002] claim that one advantage of PLTLSTR and FLTL over PLTLMIN and PLTLSIM is that the former perform a dynamic analysis of rewards capable of detecting irrelevance of variables to particular policies, e.g. to the optimal policy. Our experiments confirm this claim. However, as for reachability, whether the goal or the triggering condition in a guard formula becomes irrelevant plays an important role in determining whether a PLTLSTR or FLTL approach should be taken: PLTLSTR is able to dynamically ignore the goal, while FLTL is able to dynamically ignore the trigger.

This is illustrated in Figures 8 and 9. In both figures, the domain considered is ON/OFF with $n = 6$ propositions, the guard formula is $g \wedge \ominus^n c$ as before, here with both $g$ and $c$ achievable. This guard formula is assigned a fixed reward. To study the effect of dynamic irrelevance of the goal, in Figure 8, achievement of $\neg g$ is rewarded by the value $r$ (i.e. we have $\neg g : r$ in PLTL). In Figure 9, on the other hand, we study the effect of dynamic irrelevance of the trigger and achievement of $\neg c$ is rewarded by the value $r$. Both figures show the runtime of the methods as $r$ increases.

Achieving the goal, resp. the trigger, is made less attractive as $r$ increases up to the point where the guard formula becomes irrelevant under the optimal policy. When this happens, the run-time of PLTLSTR resp. FLTL, exhibits an abrupt but durable improvement. The figures show that FLTL is able to pick up irrelevance of the trigger, while PLTLSTR is able to exploit irrelevance of the goal. As expected, PLTLMIN whose analysis is static does not pick up either and performs consistently badly.

## 5 CONCLUSION AND FUTURE WORK

NMRDPP proved a useful tool in the experimental analysis of approaches for decision processes with Non-Markovian rewards. Both the system and the analysis are the first of their kind. We were able to identify a number of general trends in the behaviours of the methods and to provide advice concerning which are best suited to certain circumstances. We found PLTLSTR and FLTL preferable to state-based PLTL approaches in most cases. If one insists on using the latter, we strongly recommend preprocessing. In all cases, attention should be paid to the syntax of the reward formulae and in particular to minimising its length. FLTL is the technique of choice when the reward requires tracking a long sequence of events or when the desired behaviour is composed of many elements with identical rewards. For guard formulae, we advise the use of PLTLSTR if the probability of reaching the goal is low or achieving it is very risky, and conversely, of FLTL if the probability of reaching the triggering condition is low or if reaching it is very risky. For obvious reasons, this first report has focused on artificial domains. It remains to be seen what form these results take in the context of domains of more practical interest. Another topic for future work is to exploit our findings to design improved NMRDP solution methods.

**Acknowledgements** Thanks to John Slaney for useful discussions.

## References

[Bacchus et al., 1996] F. Bacchus, C. Boutilier, and A. Grove. Rewarding behaviors. In *Proc. AAAI-96*, pages 1160–1167, 1996.

[Bacchus et al., 1997] F. Bacchus, C. Boutilier, and A. Grove. Structured solution methods for non-markovian decision processes. In *Proc. AAAI-97*, pages 112–117, 1997.

[Bacchus and Kabanza, 2000] F. Bacchus and F. Kabanza. Using temporal logic to express search control knowledge for planning. *Artificial Intelligence*, 116(1-2), 2000.

[Bonet and Geffner, 2003] B. Bonet and H. Geffner. Labeled RTDP: Improving the convergence of real-time dynamic programming. In *Proc. ICAPS-03*, 2003.

[Boutilier et al., 1995] C. Boutilier, R. Dearden, and M. Goldszmidt. Exploiting structure in policy construction. In *Proc. IJCAI-95*, pages 1104–1111, 1995.

[Feng and Hansen, 2002] Z. Feng and E. Hansen. Symbolic LAO* search for factored markov decision processes. In *Proc. AAAI-02*, 2002.

[Haddawy and Hanks, 1992] P. Haddawy and S. Hanks. Representations for decision-theoretic planning: Utility functions and deadline goals. In *Proc. KR-92*, pages 71–82, 1992.

[Hansen and Zilberstein, 2001] E. Hansen and S. Zilberstein. LAO*: A heuristic search algorithm that finds solutions with loops. *Artificial Intelligence*, 129:35–62, 2001.

[Hoey et al., 1999] J. Hoey, R. St-Aubin, A. Hu, and C. Boutilier. SPUDD: stochastic planning using decision diagrams. In *Proc. UAI-99*, 1999. SPUDD is available from http://www.cs.ubc.ca/spider/staubin/Spudd/.

[Howard, 1960] R.A. Howard. *Dynamic Programming and Markov Processes*. MIT Press, Cambridge, MA, 1960.

[Thiébaux et al., 2002] S. Thiébaux, F. Kabanza, and J. Slaney. Anytime state-based solution methods for decision processes with non-markovian rewards. In *Proc. UAI-02*, pages 501–510, 2002.